\documentclass[letterpaper]{article} 
\usepackage[submission]{aaai25}  
\usepackage{times}  
\usepackage{helvet}  
\usepackage{courier}  
\usepackage[hyphens]{url}  
\usepackage{graphicx} 
\usepackage{natbib}  
\usepackage{caption} 
\usepackage{algorithm}
\usepackage{amsmath}  
\usepackage{algpseudocode}  

\usepackage{newfloat}
\usepackage{listings}
\DeclareCaptionStyle{ruled}{labelfont=normalfont,labelsep=colon,strut=off} 
\floatstyle{ruled}
\newfloat{listing}{tb}{lst}{}
\floatname{listing}{Listing}
\usepackage{natbib}
\usepackage[utf8]{inputenc} 
\usepackage[T1]{fontenc}    
\usepackage{url}            
\usepackage{booktabs}       
\usepackage{amsfonts}       
\usepackage{nicefrac}       
\usepackage{microtype}      
\usepackage{xcolor}         
\usepackage{longtable}
\usepackage{graphicx}
\usepackage{array}
\usepackage{multirow}
\usepackage{graphicx} 
\usepackage{amssymb}  
\usepackage{xcolor,colortbl}
\usepackage{xspace}
\usepackage{makecell}
\definecolor{darkred}{rgb}{0.6,0,0}
\usepackage[breaklinks=true,
            colorlinks,
            linkcolor = darkred,
            urlcolor  = magenta, 
            citecolor = teal,
            bookmarks = false]{hyperref}
\usepackage{boldline}
\usepackage{pifont}
\usepackage{tabularx}
\usepackage{graphicx}
\usepackage[export]{adjustbox}
\usepackage{fancyvrb}
\usepackage{fvextra}
\usepackage[most,breakable]{tcolorbox}
\usepackage{url}        
\usepackage{caption}
\usepackage{pgfplots}
\pgfplotsset{compat=newest}
\usepackage{url}
\usepgfplotslibrary{units}

\newcommand{\ours}[0]{\textsc{HiAgent}\xspace}
\newcommand{\std}[0]{\textsc{Standard}\xspace}

\def\showauthors@on{T} 

\title{\ours: Hierarchical Working Memory Management for Solving Long-Horizon Agent Tasks with Large Language Model}
\author{
    Mengkang Hu\textsuperscript{\rm 1},
    Tianxing Chen\textsuperscript{\rm 1},
    Qiguang Chen\textsuperscript{\rm 2},
    Yao Mu\textsuperscript{\rm 1},
    Wenqi Shao\textsuperscript{\rm 3},
    Ping Luo\textsuperscript{\rm 1}\equalcontrib
}   
\affiliations{
    \textsuperscript{\rm 1}The University of Hong Kong, \textsuperscript{\rm 2}Harbin Institution of Technology, \textsuperscript{\rm 3} Shanghai Artificial Intelligence Laboratory\\
    \vspace{0.5em}
    \url{https://github.com/HiAgent2024/HiAgent}
}

\begin{document}
\maketitle
\begin{abstract}
Large Language Model (LLM)-based agents exhibit significant potential across various domains, operating as interactive systems that process environmental observations to generate executable actions for target tasks.
The effectiveness of these agents is significantly influenced by their memory mechanism, which records historical experiences as sequences of action-observation pairs.
We categorize memory into two types: cross-trial memory, accumulated across multiple attempts, and in-trial memory (\textit{working memory}), accumulated within a single attempt. 
While considerable research has optimized performance through cross-trial memory, the enhancement of agent performance through improved working memory utilization remains underexplored.
Instead, existing approaches often involve directly inputting entire historical action-observation pairs into LLMs, leading to redundancy in long-horizon tasks.
Inspired by human problem-solving strategies, this paper introduces \ours, a framework that leverages subgoals as memory chunks to manage the working memory of LLM-based agents hierarchically. 
Specifically, \ours prompts LLMs to formulate subgoals before generating executable actions and enables LLMs to decide proactively to replace previous subgoals with summarized observations, retaining only the action-observation pairs relevant to the current subgoal. 
Experimental results across five long-horizon tasks demonstrate that \ours achieves a twofold increase in success rate and reduces the average number of steps required by 3.8.
Additionally, our analysis shows that \ours consistently improves performance across various steps, highlighting its robustness and generalizability.
\end{abstract}

\section{Introduction}

\begin{figure}[ht]
    \centering
    \includegraphics[width=\linewidth]{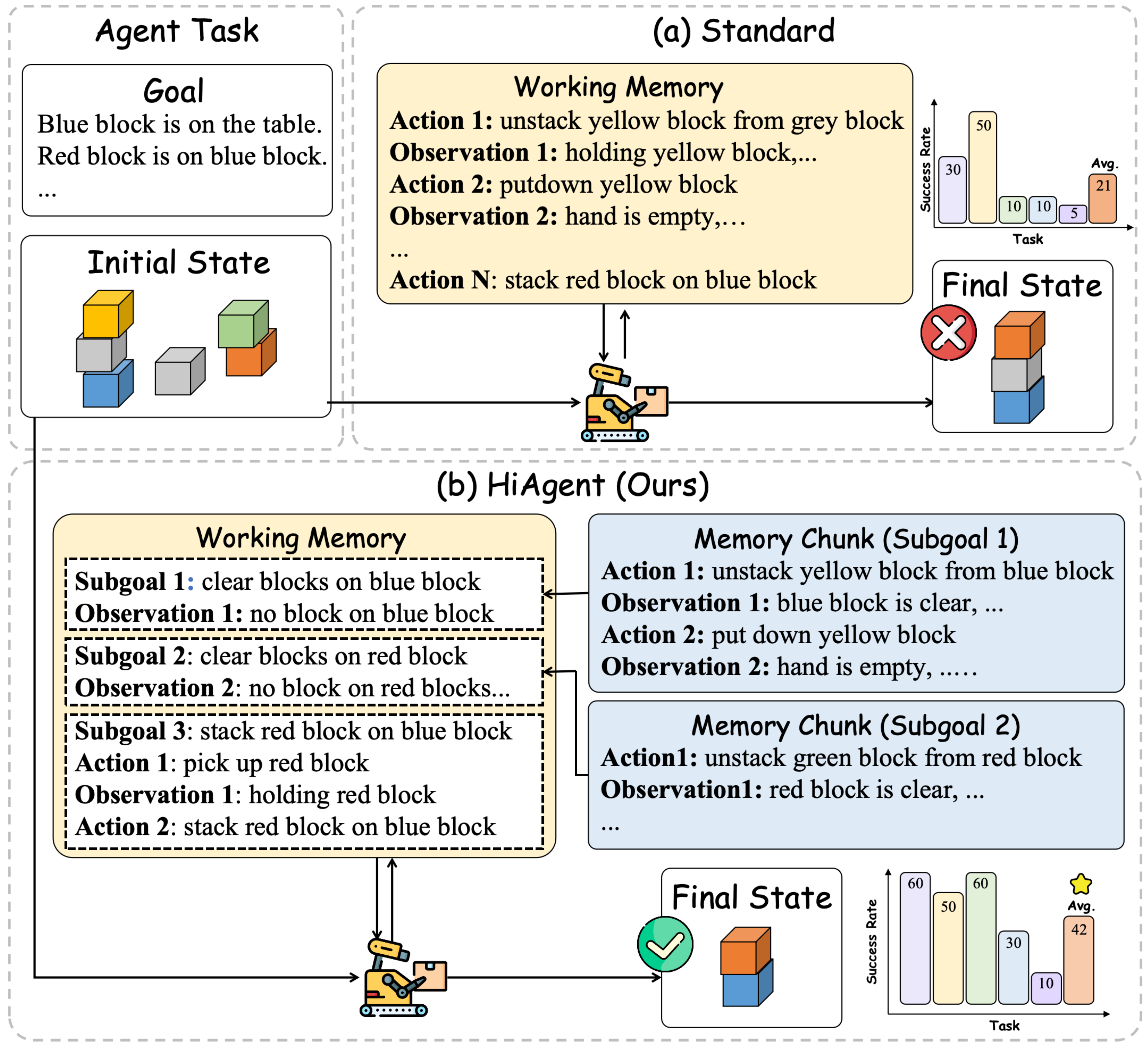}
    \caption{%
\textbf{Top right}: A commonly adopted paradigm \std for LLM-based agents includes: i) prompts LLMs to generate one action; ii) executes the generated action and then append the obtained observation to the LLM's context (working memory); and iii) generates the next action.
\textbf{Bottom}: Instead of incorporating all historical action-observation pairs into the working memory, \ours leverage subgoals as memory chunks, with a summarized observation as the observation for each memory chunk. 
\ours achieves an average success rate improvement of twofold (42 vs. 21) across five long-horizon tasks.
    }
    \vspace{-10pt}
    \label{fig:teaser}
\end{figure}

Owing to the development of powerful reasoning capabilities of Large Language Models (LLMs) in recent years~\cite{opeiai2022gpt,openai2023gpt4,meta2024llama3,touvron2023llama,jiang2023mistral}, LLM-based agents have demonstrated significant potential in various applications~\cite{xie2023openagents,wang2024survey,xi2023rise}, such as software development~\cite{hong2023metagpt,bairi2024codeplan}, robotic planning~\cite{yao2022react,puig2018virtualhome,singh2023progprompt,huang2022language}, simulating human behavior~\cite{park2023generative}, etc.
Typically, an LLM-based agent refers to an interactive system that processes environmental observations, maintains context across multiple rounds of dialogue, and outputs executable actions tailored to completing a given task.
\textit{Memory} is one of the critical components of LLM-based agents, involving how agents store and utilize past experiences.
When handling a specific task, an agent's memory can be divided into cross-trial and in-trial memory (also as known as \textit{working memory}). 
Cross-trial memory typically consists of the historical trajectory information accumulated across multiple attempts at the current task. In contrast, in-trial memory pertains to the information relevant to the current trial.
While many papers have explored leveraging cross-trial memory to optimize agent performance~\cite{shinn2024reflexion,zhao2024expel,guo2023empowering}, few have investigated ways to better utilize working memory. 
Existing LLM-based agent literature primarily employs the \std strategy illustrated in Figure~\ref{fig:teaser}, where all action-observation pairs in working memory are directly incorporated into the context when prompting LLMs~\cite{liu2023agentbench,ma2024agentboard,yao2022react}.
Although this approach transmits the historical information to the LLM as comprehensively as possible, it encounters issues in \textit{long-horizon agent tasks}. 
Such tasks typically require the agent to perform numerous actions to complete the task, resulting in an extensive working memory. 
This lengthy working memory creates a redundant context, hindering LLMs from maintaining coherent strategies and making accurate predictions over extended periods.

Drawing on principles of cognitive science~\cite{newell1972human, anderson2013architecture}, humans typically decompose a complex problem into multiple subproblems, addressing each individually. 
Each subproblem is treated as a memory ``chunk,'' thereby reducing the cognitive load on working memory~\cite{miller1956magical}. 
By focusing on the results of completed subproblems rather than their detailed execution, humans effectively manage cognitive resources and improve their efficiency in solving complex, long-horizon tasks.
Inspired by human cognition and problem-solving strategies, we propose a sophisticated hierarchical working memory management framework \textbf{\ours} tailored for long-horizon agent tasks. 
The core idea of \ours is to trigger LLMs to generate subgoals, with each subgoal serving as a chunk of the working memory.
Specifically, as shown in Figure~\ref{fig:main}, we first prompt the LLM to generate a subgoal, then create actions to achieve the subgoal and store the corresponding action-observation pairs in a memory chunk. 
Once the subgoal is completed, we summarize the memory chunk and append the subgoal-observation pair to the working memory.
In a word, \ours triggers LLMs to proactively decide to replace previous subgoals with summarized observations while retaining only the action-observation pairs relevant to the current subgoal.
To provide more flexible working memory management, we also introduce a trajectory retrieval module, which can retrieve the detailed trajectory information of specific past subgoals when necessary.

To validate the effectiveness and efficiency of \ours, we conducted experiments on five long-horizon agent tasks from AgentBoard~\cite{ma2024agentboard}. 
The experimental results show that the success rate of \ours is twice that of the \std strategy, and it exceeds the \std strategy by 23.94\% in progress rate. 
Additionally, \ours is more efficient than \std strategy, reducing the average number of steps to complete tasks by 3.8, the context length by 35.02\%, and the run time by 19.42\%.
Furthermore, to demonstrate that redundant context impairs the performance of LLM-based agents in long-horizon tasks, we compared \ours to a method that generates subgoals without disregarding the detailed trajectory information of past subgoals. 
Experimental results show that \ours improved the success rate by 20\% while reducing both runtime and the number of steps.
By analyzing model performance across varying step counts, we found that \ours not only consistently outperformed \std on progress rate but also showed a higher likelihood of generating executable actions as the number of steps increased.


\section{Preliminary}

\subsection{Large Language Model based Agent}

Large Language Model (LLM) based agents are intelligent autonomous systems designed to perform complex tasks. 
These tasks can be formalized as a partially observable Markov decision process (POMDP), characterized by the tuple $(S, O, A, T, R)$, where:
$S$ denotes the state space;
$O$ represents the observation space;
$A$ signifies the action space;
$T: S \times A \rightarrow S$ embodies the transition function;
$R: S \times A \rightarrow R$ encapsulates the reward function;
An LLM-based agent operates as a policy $\pi(a_t|I,o_t,a_{t-1},o_{t-1},\ldots,a_{0},o_{0})$, which, given the historical action-observation pairs and instructions $I$ (encompassing in-context examples, environmental descriptions, etc.), generates an executable action $a_t \in A$.
Each action precipitates a new state $s_{t+1} \in S$ and a subsequent observation $o_{t+1} \in O$. This iterative interaction persists until either task completion or the agent reaches a predetermined maximum number of steps.

\subsection{Working Memory}

From the cognitive science perspective, working memory enables individuals to hold and manipulate information in real-time, facilitating complex cognitive tasks such as reasoning, comprehension, and learning~\cite{newell1972human,anderson2013architecture}.
In LLM-based agents, we define working memory as the essential historical information required by the LLM at a given moment $t$ to complete the current task.
Effective working memory management allows for better integrating past experiences and current stimuli, leading to more informed and accurate decisions. 
It can be likened to the human process of attentional control and cognitive updating, which involves selectively focusing on relevant information, filtering out distractions, and continually updating the mental workspace with new and pertinent data. 
The \std approach in Figure~\ref{fig:teaser} stores all historical action-observation pairs in working memory, i.e., $m_{t}^{std}=(o_t,a_{t-1},o_{t-1},\ldots,a_{0},o_{0})$. Although this provides the LLM with comprehensive information, it also introduces redundancy, complicating the LLM's processing.
\section{Methodology}
\label{sec:methodology}
\vspace{-5pt}

\begin{figure}
\centering
\includegraphics[width=\linewidth]{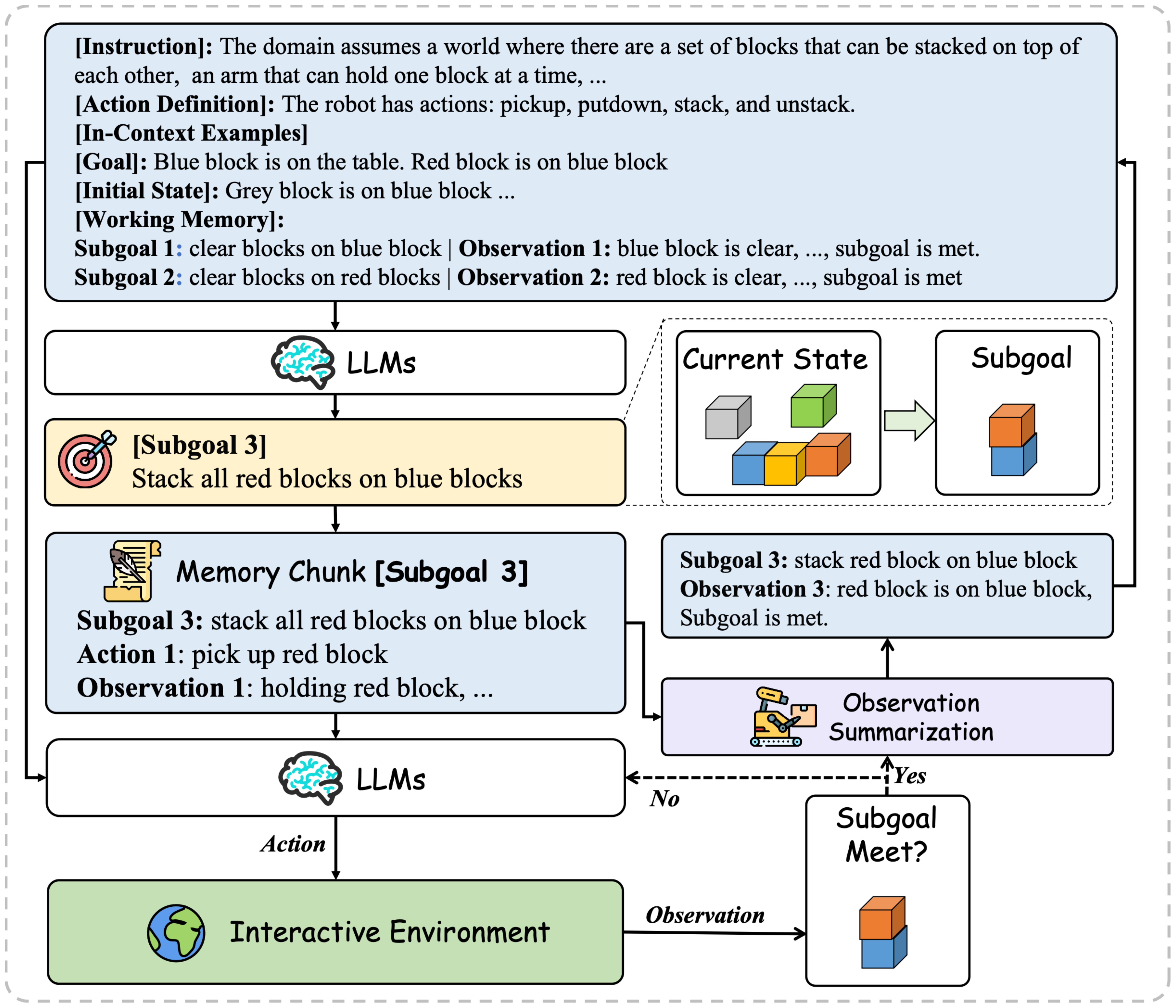}
\caption{An overview of the process of \ours.}
\label{fig:main}
\end{figure}

\subsection{Overview}

The core idea of \ours is to employ subgoals for hierarchical management of working memory. 
More specifically, as is shown in Figure~\ref{fig:main}, the process of \ours can be described as follows: 
\textbf{(1)} Before generating specific grounded actions, we prompt the LLM to first formulate a subgoal $g_i$. Each subgoal serves as a milestone within the overall task. 
\textbf{(2)} Subsequently, the LLM generates precise actions to accomplish this subgoal. 
\textbf{(3)} Upon the LLM's determination that a particular subgoal has been fulfilled, we synthesize the corresponding action-observation pairs into a summarized observation $s_i$ (\S\ref{sec:summarization}).
We then obscure the action-observation pairs within the context, substituting them with $s_i$. 
Consequently, the working memory of \ours can be formalized as $m_t=(g_0,s_0,....,g_{n-1},s_{n-1},g_n,a_{n0},o_{n1},...)$.
\textbf{(4)} Additionally, we have incorporated a retrieval module to facilitate more flexible memory management(\S\ref{sec:retrieval}). 
For instance, if the $q_{th}$ subgoal is retrieved, we input the detailed action-observation pairs into the context rather than the summarized observation, i.e.,  $m'_t=(g_0,s_0,....,g_q,a_{q0},a_{q0},...,g_n,a_{n0},o_{n0},...)$.

\subsection{Subgoal-based Hierarchical Working Memory}
\label{sec:subgoal}

As is shown in Figure~\ref{fig:main}, at each time step, the LLM can either generate the next action for the current subgoal or generate a new subgoal when it determines that the existing subgoal has been accomplished. 
For the current subgoal, the agent retains all action-observation pairs, providing a detailed context for immediate decision-making. For past subgoals, only a summarized version of the observations is kept. 
This subgoal-based hierarchical management approach in \ours is deeply motivated by cognitive science principles, drawing parallels with human cognition and problem-solving strategies~\cite{newell1972human,anderson2013architecture}. 
Employing subgoals to compartmentalize action-observation pairs can be conceptualized as a form of chunking methodology. In human cognition, chunking allows individuals to group related information into meaningful units, thereby overcoming working memory limitations~\cite{miller1956magical}. Similarly, \ours utilizes subgoals as cognitive chunks, encapsulating related actions and observations. This chunking mechanism enables the system to handle complex sequences of information more effectively, reducing cognitive load and enhancing overall performance.
Furthermore, 
by generating subgoals before specific actions, the system mimics the human tendency to break down larger objectives into more manageable components. 
This methodology enhances computational efficiency and aligns with established theories of human information processing.

\subsection{Observation Summarization}
\label{sec:summarization}

The process of observation summarization can be formalized as $s_i = S(g_i, o_0, a_0, ..., o_t)$, where $S$ can be implemented using either a Large Language Model (LLM) or alternative text summarization models.
This function encapsulates the synthesis of historical observations and actions, contextualized by the current subgoal, to produce a concise representation of the agent's state.
Furthermore, a crucial component of the summarized observation is assessing whether the current subgoal has been achieved. This evaluation serves as a pivotal guide for future subgoal generation, facilitating adaptive and goal-oriented behavior in the agent's decision-making process.
By doing so, the agent can maintain a condensed yet informative context, balancing the need for historical information with efficiency. The example prompt is as follows:

\texttt{\scriptsize
\noindent You are an advanced AI system tasked with summarizing and analyzing a series of action-observation pairs (trajectories) and determining whether a specific subgoal has been met. \\
Your goal is to create a summary that captures all essential information, decisions, and outcomes from the given trajectories, and indicate whether the subgoal has been met based on the summarized observations.\\
If there are no valid actions taken, you need to analyze the reason.\vspace{5px}\\
\#\#\# Instructions:\\
1. Provide a summarized observation related to the subgoal in a concise manner.\\
2. Determine whether the subgoal has been met.\\
3. Do not output anything except whether summary and subgoal are met. Your output should be only one line. Do not output things like `\#\#Summary', `\#\#Summary and Analysis'.\\\\
\{example\}\vspace{5px}\\
\#\#Trajectory\\
\{formatted\_trajectory\}\vspace{5px}\\
\#\#Subgoal:\\
\{subgoal\}\vspace{5px}\\
\#\#\#Output:\vspace{5px}
}

\subsection{Trajectory Retrieval}
\label{sec:retrieval}
Despite the summarization, there may be instances where detailed past trajectory information becomes crucial for immediate decision-making. 
For instance, when a past subgoal execution fails, we need detailed trajectory information to determine the cause of failure. Moreover, reviewing past successful experiences can also increase the likelihood of success when facing novel challenges and scenarios.
To address this, we introduce a trajectory retrieval module. 
To address this, we introduce a trajectory retrieval module. When the LLM determines that detailed information from a past subgoal is necessary, it generates a retrieval function to recall the complete action-observation pairs for that subgoal, analogous to the way to generate actions.
This selective retrieval allows the agent to access detailed historical data on-demand without consistently carrying the full context.
\section{Experiments}
\definecolor{mygray}{gray}{0.9}
\definecolor{newgreen}{RGB}{78, 173, 102}

\newcommand{\worse}[2]{{#1 \textcolor{purple}{\small{-#2}}}}
\newcommand{\better}[2]{{#1  \textcolor{newgreen}{\small{+#2}}}}
\newcommand{\bfbetter}[2]{\cellcolor{white}{\textbf{#1}  \textcolor{newgreen}{\small{+#2}}}}
\newcommand{\bfworse}[2]{\cellcolor{white}{\textbf{#1}  \textcolor{newgreen}{\small{-#2}}}}
\newcommand{\upbetter}[1]{\cellcolor{white}{\textbf{#1}}}
\newcommand{\ccbf}[1]{\textbf{#1}}

\begin{table*}[htbp]
  \centering
    \caption{Performance of \std and \ours on 5 long-horizon agent tasks. We report on four metrics: Success Rate (\textbf{SR}), Progress Rate (\textbf{PR}), Average Steps (\textbf{Steps}), and Context Efficiency (\textbf{Context}), Run Time (\textbf{Time}). The symbol $\uparrow$ indicates that a higher value for the metric is preferable, while $\downarrow$ signifies that a lower value is considered better. In the \textbf{\textit{Overall}} section, the result is obtained by averaging the values of a certain metric across various tasks.}
    \begin{tabular}{llllll}
    \toprule
          & \textbf{SR} $\uparrow$ & \textbf{PR} $\uparrow$ & \textbf{Steps} $\downarrow$ & \textbf{Context} $\downarrow$ & \textbf{Time} $\downarrow$  \\
    \midrule
    \multicolumn{5}{l}{\textbf{\textit{{Blocksworld}}}} \\
    {\std} & {30.00} & {35.00} & {25.00} & {100\%} & {100\%}  \\
         
    {\ours} & \bfbetter{60.00}{30.00} & \bfbetter{80.00}{45.00} & \bfworse{18.60}{6.40} & \bfworse{67.46\%}{32.54\%} & \bfworse{63.47\%}{36.53\%}\\

    \midrule
    \multicolumn{5}{l}{\textbf{\textit{{Gripper}}}} \\
    {\std} & \upbetter{50.00} & \upbetter{87.75} & {25.20} & {100\%} & {100\%} \\
    {\ours} & \bfbetter{50.00}{0.00} & \worse{86.25}{1.50} & \bfworse{24.80}{0.40} & \bfworse{49.99\%}{50.01\%} & \bfworse{70.46\%}{29.54\%} \\

    \midrule
    \multicolumn{5}{l}{\textbf{\textit{{Tyreworld}}}} \\
    {\std} & {10.00} & {39.28} & {28.40} & {100\%} & {100\%} \\
    {\ours} & \bfbetter{60.00}{50.00} & \bfbetter{75.83}{36.55} & \bfworse{19.00}{9.4} & \bfworse{73.58\%}{26.42\%} & \bfworse{77.58\%}{22.42\%} \\
    
    \midrule
    \multicolumn{5}{l}{\textbf{\textit{{Barman}}}} \\
    {\std} & {10.00} & {17.50} & {26.85} & {100\%} & {100\%} \\
    {\ours} & \bfbetter{30.00}{20.00} & \bfbetter{40.83}{23.33} & \bfworse{24.5}{2.35} & \bfworse{67.02\%}{32.98\%} & \bfworse{95.54\%}{4.46\%} \\ 

    \midrule
    \multicolumn{5}{l}{\textbf{\textit{{Jericho}}}} \\
    {\std} & {5.00} & {13.51} & {26.60} & {100\%} & {100\%} \\
    {\ours} & \bfbetter{10.00}{5.00} & \bfbetter{29.85}{16.34} & \bfworse{26.15}{0.45} & \bfworse{66.86\%}{33.14\%} & \bfworse{95.85\%}{4.15\%} \\
    \hline
    
    \midrule
    \multicolumn{5}{l}{\textbf{\textit{Overall}}} \\
    {\std} & {21.00} & {38.61} & {26.41} & {100\%} & {100\%} \\
    {\ours} & \bfbetter{42.00}{21.00} & \bfbetter{62.55}{23.94} & \bfworse{22.61}{3.80} & \bfworse{64.98\%}{35.02\%} & \bfworse{80.58\%}{19.42\%} \\
    \bottomrule
    \end{tabular}%
  \label{tab:main-results}%
\end{table*}

\subsection{Experimental Setup}

\paragraph{Evaluation Tasks}
We conduct the experiments on five long-horizon agent tasks, which typically require more than 20 steps: 
\textit{(i)} \textbf{Blocksworld} requires the model to arrange the blocks into a specified target configuration by executing a series of moves;
\textit{(ii)} \textbf{Gripper} involves moving objects between different rooms;
\textit{(iii)} \textbf{Tyreworld} simulates changing a car tire, including removing the flat tire, replacing it with a spare, and installing the new tire;
\textit{(iv)} \textbf{Barman} emulates a bartender's tasks in mixing cocktails, including combining various ingredients, shakers, and garnishing drinks;
\textit{(v)} \textbf{Jericho}~\cite{hausknecht2020jericho} is a suite of text-based adventure game environments designed to evaluate agents' ability to navigate and interact with fictional worlds. 
More details can be found in Appendix~\ref{app:task_detail}.

\paragraph{Evaluation Metrics}
We use multiple metrics to evaluate both the effectiveness and efficiency of LLM-based agents in solving long-horizon tasks:
\textit{(i)} \textbf{Progress Rate}~\cite{ma2024agentboard} evaluates the advancement toward task completion. Specifically, a task consists of multiple goal conditions, and the progress rate is the proportion of goal conditions fulfilled by the model out of the total number of goal conditions.
\textit{(ii)} \textbf{Success Rate} measures the percentage of successful task completions. The success rate is 1 when the progress rate is 1.
\textit{(iii)} \textbf{Average Steps} counts the steps taken to complete the task;
\textit{(iv)} \textbf{Context Efficiency} is defined as the mean number of tokens in the in-trial context across all steps required to complete a given task. 
\textit{(v)} \textbf{Run Time} evaluates the time required to complete tasks.

\paragraph{Baselines}
\std prompting strategy is a predominantly used method in current LLM-based agent literature~\cite {yao2022react,ma2024agentboard,liu2023agentbench}. 
It operates by taking one action followed by one observation, providing a comparative baseline for evaluating the performance of \ours.

\paragraph{Implementation Details}
The implementation of evaluation tasks is based on AgentBoard~\cite{ma2024agentboard}.
We set a maximum step limit of 30 for task configuration and provide one in-context example for each task.
We employ GPT-4 (gpt-4-turbo)\footnote{We utilized the model via OpenAI API service.} as the LLM backbone for our experiments, serving both as the agent policy and the observation summarization model. 
We set the \textit{temperature} hyperparameter for LLM inference to 0 and \textit{topp} to 1.
Detailed prompt examples are provided in the Appendix~\ref{appendix:prompt_examples}.

\subsection{Main Results}

As shown in Table~\ref{tab:main-results}, \ours demonstrated substantial advancements over \std. Overall, in terms of effectiveness, it increased the success rate by 21\% and the progress rate by 23.94\%. Regarding task execution efficiency, it reduced the average number of steps to completion by 3.8, decreased the number of context tokens consumed by 35\%, and reduced the run time by 19.42\%. Furthermore, in certain tasks (blocksworld, barman, jericho), \ours even achieved more than double the progress rate improvement while maintaining efficiency. In tyreworld, the model not only achieved a 50\% improvement in success rate but also reduced the average number of steps by 9.4. Although the progress rate slightly decreased by 1.5\% in the gripper task, context token usage was reduced by over 50\%.

\noindent We can draw several conclusions from previous discussions: 

\noindent \textbf{(1)} \ours is more \textbf{effective} than \std, achieving huge improvements on both success rate and progress rate.

\noindent \textbf{(2)} \ours is also more \textbf{efficient} than \std, requiring fewer steps to complete tasks, utilizing shorter context lengths, and achieving faster runtime.




\section{Analysis}

To gain deeper insights into our approach, we explored the following research questions:

\noindent (1) \textit{Are all modules effective for \ours?}

\noindent (2) \textit{Is \ours consistently superior to the baseline at different steps?}


\noindent (3) \textit{Is improvement of \ours solely derived from task decomposition?} 

\noindent (4) \textit{How effective are the frameworks in generating executable actions?} 

\noindent (5) \textit{Are the observed performance improvements in \ours statistically significant compared to \std?}


\subsection{Answer 1: All Modules in \ours are Effective for \ours}

In this section, we conducted albation study to explore whether \textit{Observation Summarization} and \textit{Trajectory Retrieval} are effective. 

\paragraph{Observation Summarization is effective.} 
We heuristically use the observation corresponding to the last action as the summarized observation when removing the \textit{Observation Summarization} module. As is shown in Table~\ref{tab:ablation} (``\textit{w/o OS}''), there is a significant decline in performance across all metrics. 
Specifically, the success rate and progress rate were significantly impacted, decreasing by 30\% and 7.6\%, respectively.
It indicates that the observation summarization module can comprehensively aggregate the detailed information within a trajectory, thereby aiding the reasoning of an LLM-based agent.

\paragraph{Trajectory Retrieval is also crucial for performance enhancement.}
We hide all the detailed trajectory information of previous subgoals at each time step to verify the effectiveness of \textit{Trajectory Retrieval}.
According to the results in Table~\ref{tab:ablation} (``\textit{w/o TR}''), the success rate decreased by 10\%, and the average steps increased by 1.2. 
This is because, while trajectory retrieval lengthens the reasoning steps of the LLM, it allows the agent to flexibly retrieve past trajectories under certain subgoals, which is more beneficial for identifying errors in previous actions.

\paragraph{The combination of Observation Summarization and Trajectory Retrieval yields significant improvement.}
We conducted an experiment where both modules were removed to validate the functionality and effectiveness of the combined \textit{Observation Summarization} and \textit{Trajectory Retrieval} modules. 
As shown in Table~\ref{tab:ablation} (``\textit{w/o OS \& TR}''), there is a noticeable performance decline compared to \ours, with the success rate decreasing by 20\%.
This decline is also evident when compared to the individual ablations of the \textit{Observation Summarization} and \textit{Trajectory Retrieval} modules, highlighting a substantial reduction in progress rate in their absence.

\begin{table*}[thbp]
  \centering
    \caption{Ablation study of \ours on tyreworld. ``\textit{w/o OS}'' refers to removing the \textit{\textbf{O}bservation \textbf{S}ummarization} module introduced by Section~\ref{sec:summarization}. ``\textit{w/o TR}'' refers to removing the \textit{\textbf{T}rajectory \textbf{R}etrieval} module introduced by Section~\ref{sec:retrieval}. ``\textit{w/o TR \& OS}'' refers to removing both modules.}
    \begin{tabular}{llllllll}
    \toprule
          \textbf{Model} & \textbf{SR} $\uparrow$ & \textbf{PR} $\uparrow$ & \textbf{Steps} $\downarrow$ & \textbf{Context} $\downarrow$ & \textbf{Time} $\downarrow$  \\
    \midrule
    {\ours} & \textbf{60.0} & 75.8 & \textbf{19.0} & \textbf{100.0\%} & \textbf{100.0\%}  \\
    \midrule  
    \textit{w/o OS} & 30.0 \textcolor{purple}{-30.0} & 68.2 \textcolor{purple}{-7.6}  & 24.2 \textcolor{purple}{+5.2} & 110.8\% \textcolor{purple}{+10.8\%} & 122.5\% \textcolor{purple}{+22.5\%} \\
    \textit{w/o TR} & 50.0 \textcolor{purple}{-10.0} & \textbf{76.9} \textcolor{newgreen}{+1.1} & 21.2 \textcolor{purple}{+2.2} & 105.0\% \textcolor{purple}{+5.0\%}& 107.5\% \textcolor{purple}{+7.5\%}\\
    \textit{w/o OS} \& TR & 30.0 \textcolor{purple}{-30.0}  & 62.4  \textcolor{purple}{-13.4} & 26.2 \textcolor{purple}{+7.2} & 107.2\% \textcolor{purple}{+7.2\%} & 121.2\% \textcolor{purple}{+21.2\%} \\
    \bottomrule
    \end{tabular}%
  \label{tab:ablation}%
\end{table*}

\subsection{Answer 2: \ours is consistently superior to \std at different steps}


\begin{figure*}[h]
    \centering
    \begin{minipage}{0.74\textwidth}
        \centering
        \includegraphics[width=\textwidth]{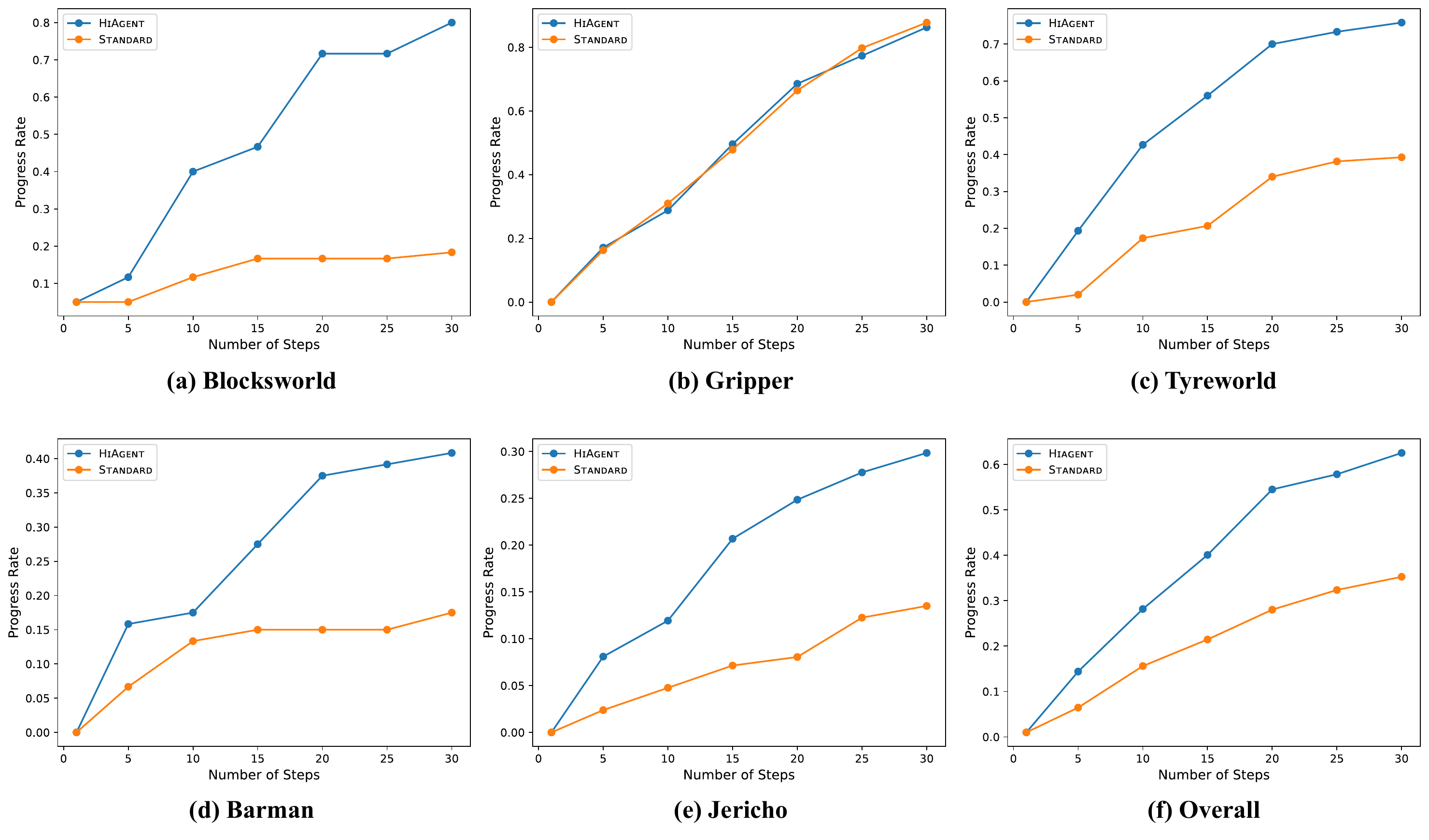}
        \caption{Progress rate at different steps.}
        \label{fig:results_steps}
    \end{minipage}
    \hfill
    \begin{minipage}{0.24\textwidth}
        \centering
        \includegraphics[width=\textwidth]{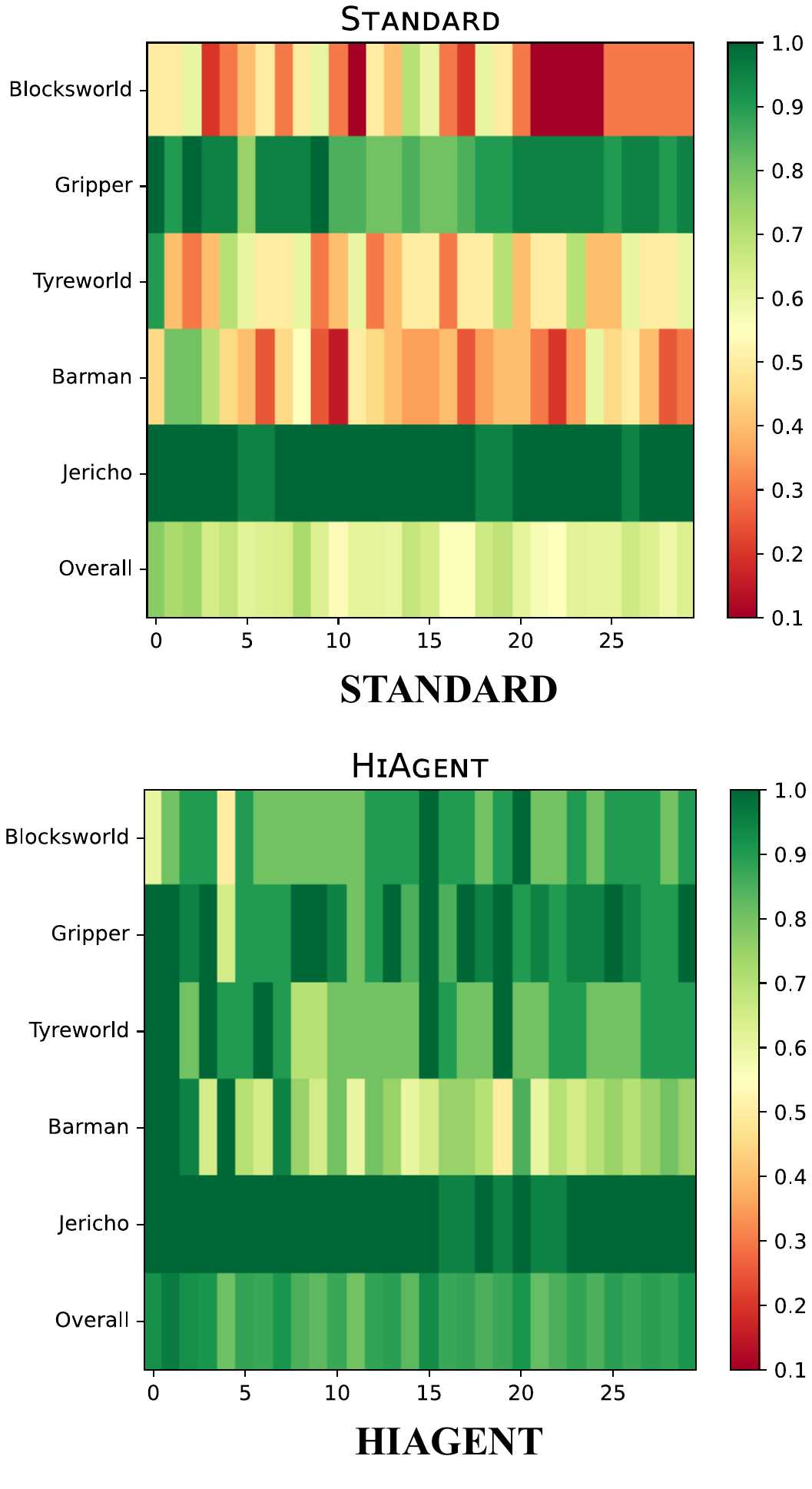}
        \caption{Executability of actions at different steps.}
        \label{fig:heatmap}
    \end{minipage}
\end{figure*}

To conduct a more granular study of \ours's performance, we present the progress rate at different step counts (in intervals of 5 steps) in Figure~\ref{fig:results_steps}. 
The experimental results indicate that overall, \ours consistently achieves a higher progress rate at each step than \std \textit{(f)}. Additionally, it is noteworthy that \ours benefits more from an increased number of steps, whereas \std does not. For example, in the blocksworld task \textit{(a)} and barman task \textit{(b)}, \std shows no progress rate increase between steps 15-25, whereas \ours exhibits continuous growth. This further demonstrates \ours's advantage in handling long-horizon agent tasks.


\subsection{Answer 3: The improvement in \ours is not solely attributed to task decomposition}
\label{sec:subgoal_analysis}

\begin{table*}[htbp]
  \centering
    \caption{
    Experimental results on tyreworld. ``w. \textsc{TD}'' refers to \textit{Task Decomposition}, i.e., having the LLM generate subgoals without concealing detailed trajectory information of previous subgoals.
    }
    \begin{tabular}{llllll}
    \toprule
          \textbf{Model} & \textbf{SR} $\uparrow$ & \textbf{PR} $\uparrow$ & \textbf{Steps} $\downarrow$ & \textbf{Context} $\downarrow$ & \textbf{Time} $\downarrow$  \\
    \midrule
    \std & 10.0 & 39.3 & 28.4 & 100\% & 100\% \\
    w. \textsc{TD} & \better{40.0}{30.0} & \better{67.4}{28.1} & 22.8 \textcolor{newgreen}{-5.6} & 112.8\% \textcolor{purple}{+12.8\%} & 105.7\% \textcolor{purple}{+5.7\%} \\
    w. \ours & \bfbetter{60.0}{50.0} & \bfbetter{75.8}{36.5} & \bfworse{19.0}{9.4} & \textbf{73.6\%} \textcolor{newgreen}{-26.4\%} & \textbf{77.6\%} \textcolor{newgreen}{-22.4\%}  \\
    \bottomrule
    \end{tabular}%
  \label{tab:decomposition}%
\end{table*}

Using LLMs to generate subgoals has been employed in numerous studies and has demonstrated considerable performance advantages~\cite{zhou2022least,yin2023lumos}. Therefore, a pertinent question arises: ``\textit{Is the performance improvement attributed to \ours merely related to task decomposition, rather than efficient working memory management?}''
To address this question, we implemented a new method that prompts the LLM to generate a subgoal before generating executable actions, followed by generating actions to achieve this subgoal. 
Unlike \ours, this approach does not obscure the detailed trajectory information of previous subgoals.
The experimental results, detailed in Table \ref{tab:decomposition}, indicate that although task decomposition can lead to a performance improvement (30\% in success rate), the success rate is still 20\% lower than \ours.
Additionally, solely using task decomposition introduces inefficiencies, increasing runtime by 5.7\% and context length by 12.8\%.
In summary, \ours is more efficient and effective than task decomposition alone.

\subsection{Answer 4: \ours is effective in generating executable actions even under long steps}


LLM-based agents sometimes generate actions that cannot be executed, such as attempting to retrieve objects from a closed container. This is typically due to LLMs' poor reasoning abilities. 
To investigate this, we calculated the proportion of executable actions generated by the model at each timestep, referred to as \textit{executability}. 
As shown in Figure~\ref{fig:heatmap}, \ours is more likely to generate executable actions than \std, further demonstrating the effectiveness of \ours. 
Additionally, we observed that \std is more prone to generating non-executable actions when the steps are longer (e.g., in the blocksworld, when the steps exceed 20, executability drops below 10\%). 
This is because, as the working memory increases, the ability of LLMs to generate executable actions decreases. In contrast, \ours maintains over 80\% executability even with longer steps, indicating that the robustness to long steps is a key factor in the strong performance on long-horizon tasks.

\subsection{Answer 5: The observed performance improvements in \ours are statistically significant compared to \std}

To validate the statistical significance of the improvements in both effectiveness and efficiency, we selected the \textit{Progress Rate} and \textit{Average Steps} metrics for analysis. 
We employed the Wilcoxon signed-rank test~\cite{woolson2005wilcoxon} for this purpose due to its suitability for comparing paired samples. 
This non-parametric test helps assess whether the observed differences are likely due to chance or represent a genuine effect.
The results of our analysis are as follows: 
\textit{(i)} For the Progress Rate, the test statistic is $144.0$ with a p-value of $2.38 \times 10^{-5}$, indicating a statistically significant difference between \ours and \std; 
\textit{(ii)} For the Average Steps, the test statistic is $112.5$ with a p-value of $0.0016$, also demonstrating a statistically significant difference.
These results confirm that the observed improvements in both effectiveness and efficiency are not due to random variation, underscoring the superiority of \ours.
\section{Related Work}
\label{sec:related}

\paragraph{Large Language Model based Agent.}

Large Language Models (LLMs) have revolutionized the field of language agents, endowing them with the prowess to tackle intricate challenges through a logical sequence of actions~\cite{xie2023openagents,hong2023metagpt,xi2023rise,wang2024survey,yao2022react,zhou2023language}. 
A series of works explored various applications of LLM-based agents, such as code generation~\cite{wang2023mint,lin2018nl2bash}, web browsing~\cite{yao2022webshop,zhou2023webarena,pan2024webcanvas,li2024websuite}, robotics~\cite{chevalier2018babyai,shridhar2020alfworld,mu2024robocodex,mu2024embodiedgpt}, tool use~\cite{li2023api,wu2024seal,qin2023toolllm}, reasoning~\cite{yang2024aqa}, planning~\cite{xie2024travelplanner}, conducting research~\cite{kang2024researcharena}, chip design and more.
Additionally, lots of works explored the application of LLM-based agents in the field of multi-agent systems~\cite{hong2023metagpt,zhang2023building,wu2023autogen,li2023camel,chen2023agentverse}.
This paper introduces a working memory management framework \ours that can be universally applied to enhance the performance of other agent frameworks. 
For example, ReAct~\cite{yao2022react} introduces a method where the LLM generates a chain of thought~\cite{wei2022chain} before generating actions, and the trajectory formed by the triplet of \textit{``(thought, action, observation)''} can be managed using \ours. 
Additionally, \ours has the potential to alleviate information management challenges in multi-agent frameworks~\cite{hong2023metagpt}.

\paragraph{Planning.}
Planning is a cornerstone of human intelligence, representing a systematic approach to achieving goals through a series of deliberate actions~\cite{yao2024tree,zhang2023reasoning,xu2023rewoo,song2023llmplanner,wang2023describe,huang2023grounded,huang2022inner,liu2023llm+,guan2023leveraging,zhao2024large,ruan2023tptu,aghzal2023can}. It involves breaking down complex tasks into manageable sub-tasks, searching for potential solutions, and achieving a desired goal.
%
This cognitive ability is fundamental to human-level intelligence and has been a focal point of research in various domains, including robotics~\cite{hu2023tree,huang2022language,singh2023progprompt,brohan2023can,valmeekam2024planbench,puig2018virtualhome}, travel planning~\cite{xie2024travelplanner}, warehouse-level coding~\cite{bairi2024codeplan}, tool use~\cite{liu2024tool} and so on.
Least-to-most~\cite{zhou2022least} and Plan-and-solve~\cite{wang2023plan} propose decomposing a complex question into a series of sub-questions. However, when answering each sub-question, it inputs all previous answers into the LLM, leading to context inefficiency.
Lumos~\cite{yin2023lumos} and XAgent~\cite{xagent2023} introduce an independent planning module for generating subgoals and use full context in the grounding module to complete each subgoal.
\ours distinguishes itself from the literature by not only utilizing planning to enhance task performance but also by using subgoals as memory chunks to manage working memory hierarchically. 
This approach brings context efficiency and surpasses methods that rely solely on planning, as discussed in Section~\ref{sec:subgoal_analysis}.

\paragraph{Memory.}
The memory module in LLM-based agents is analogous to the human memory system, which is responsible for encoding, storing, and retrieving information~\cite{zhang2024survey}. 
The memory modules are typically divided into long-term memory and short-term memory. Long-term memory can usually be stored in an external database, while short-term memory (also known as working memory) is typically used directly as the context input of LLMs. 
Most current research papers primarily focus on managing long-term memory~\cite{alonso2024toward,maharana2024evaluating,chen2024compress,xiao2024infllm,yuan2023evolving,wang2023jarvis,majumder2023clin,hu2023chatdb,hao2024toolkengpt,lanchantin2024learning,tu2023chatlog,liang2023unleashing,kagaya2024rap}. 
Pioneer works include Memorybank~\cite{zhong2024memorybank}, with its global-level summaries, has made significant strides in distilling conversations into coherent narratives.
Other works, such as Think-in-memory~\cite{liu2023think} and the Retroformer~\cite{yao2023retroformer}, incorporated summary modules to manage long-term memories.
Unlike these works, our study investigates how optimizing the management of working memory can enhance agent performance.
Another line of research involves modifying the structure of transformers to enable large language models (LLMs) to process longer contexts, thereby extending their working memory capabilities~\cite{zhou2023recurrentgpt,chevalier2023adapting,bertsch2024unlimiformer,ruoss2023randomized,beltagy2020longformer,an2023eval}. However, existing research has identified that LLMs encounter attention loss issues with lengthy texts~\cite{liu2024lost}. Consequently, we believe that investigating more efficient management of working memory remains a valuable endeavor.

\section{Conclusion}
\label{sec:conclusion}
This paper proposes \ours, a hierarchical framework that utilizes subgoals to manage the working memory of Large Language Model(LLM)-based agents.
\ours aims to address the poor performance of LLM-based agents when handling long-horizon tasks. 
Experimental results from five long-horizon agent tasks demonstrate that \ours outperforms the baseline model across all tasks, with an overall success rate more than double that of the baseline model. 
Furthermore, \ours is more efficient, accomplishing tasks with fewer steps, in less runtime, and using shorter context.
We also conducted an ablation study to verify the effectiveness of the individual modules of \ours. 
A series of analysis experiments demonstrate that as the number of steps increases, \ours more effectively generates executable actions and consistently outperforms \std in terms of progress rate.
Additionally, we conducted a statistical test to validate the statistical significance of the improvements introduced by \ours.
We believe \ours is an effective and flexible framework that can be integrated into other agent frameworks.
In the future, we hope \ours can inspire more creative ideas on effectively managing the working memory of LLM-based agents.

\bibliography{bibliography}

\section*{Reproducibility Checklist}

\begin{enumerate}
    \item This paper:
    \begin{enumerate}
        \item Includes a conceptual outline and/or pseudocode description of AI methods introduced (yes)
        \item Clearly delineates statements that are opinions, hypotheses, and speculations from objective facts and results (yes)
        \item Provides well-marked pedagogical references for less-familiar readers to gain background necessary to replicate the paper (yes)
    \end{enumerate}
    \item Does this paper make theoretical contributions? (no)
    \item Does this paper rely on one or more datasets? (yes)
    \begin{enumerate}
        \item A motivation is given for why the experiments are conducted on the selected datasets (yes)
        \item All novel datasets introduced in this paper are included in a data appendix. (NA)
        \item All novel datasets introduced in this paper will be made publicly available upon publication of the paper with a license that allows free usage for research purposes. (NA)
        \item All datasets drawn from the existing literature (potentially including authors’ own previously published work) are accompanied by appropriate citations. (yes)
        \item All datasets drawn from the existing literature (potentially including authors’ own previously published work) are publicly available. (yes)
        \item All datasets that are not publicly available are described in detail, with an explanation of why publicly available alternatives are not scientifically satisfying. (NA)
    \end{enumerate}

    \item Does this paper include computational experiments? (yes)
    \begin{enumerate}
        \item Any code required for pre-processing data is included in the appendix. (yes)
        \item All source code required for conducting and analyzing the experiments is included in a code appendix. (yes)
        \item All source code required for conducting and analyzing the experiments will be made publicly available upon publication of the paper with a license that allows free usage for research purposes. (yes)
        \item All source code implementing new methods has comments detailing the implementation, with references to the paper where each step comes from. (yes)
        \item If an algorithm depends on randomness, then the method used for setting seeds is described in a way sufficient to allow replication of results. (yes)
        \item This paper specifies the computing infrastructure used for running experiments (hardware and software), including GPU/CPU models; amount of memory; operating system; names and versions of relevant software libraries and frameworks. (yes)
        \item This paper formally describes evaluation metrics used and explains the motivation for choosing these metrics. (yes)
        \item This paper states the number of algorithm runs used to compute each reported result. (yes)
        \item Analysis of experiments goes beyond single-dimensional summaries of performance (e.g., average; median) to include measures of variation, confidence, or other distributional information. (yes)
        \item The significance of any improvement or decrease in performance is judged using appropriate statistical tests (e.g., Wilcoxon signed-rank). (yes)
        \item This paper lists all final (hyper-)parameters used for each model/algorithm in the paper’s experiments. (yes)
        \item This paper states the number and range of values tried per (hyper-)parameter during the development of the paper, along with the criterion used for selecting the final parameter setting. (yes)
    \end{enumerate}
\end{enumerate}

\newpage
\appendix
\tcbset{colframe = blue!50!black, 
colback = white, 
toptitle=1mm, 
bottomtitle=1mm, 
fonttitle=\fontsize{10pt}{12pt}\selectfont,  breakable, enhanced jigsaw, finish broken={\tcbset{colframe=white, bottom=0mm, bottomrule=0mm}},before upper={\tcbset{colframe=white, toprule=0mm}},after={\tcbset{colframe=blue!40!black, toprule=1mm, bottomrule=1mm}},}

\definecolor{LightYellow}{RGB}{255, 255, 102} 

\section{More Details on Evaluation Tasks}
\label{app:task_detail}



\subsection{Blocksworld}

\textbf{Action List}

\noindent 1. pickup <block>: allows the arm to pick up a block from the table if it is clear and the arm is empty. After the pickup action, the arm will be holding the block, and the block will no longer be on the table or clear.\\
2. putdown <block>: allows the arm to put down a block on the table if it is holding a block. After the putdown action, the arm will be empty, and the block will be on the table and clear.\\
3. stack <block> <block>: allows the arm to stack a block on top of another block if the arm is holding the top block and the bottom block is clear. After the stack action, the arm will be empty, the top block will be on top of the bottom block, and the bottom block will no longer be clear.\\
4. unstack <block> <block>: allows the arm to unstack a block from on top of another block if the arm is empty and the top block is clear. After the unstack action, the arm will be holding the top block, the top block will no longer be on top of the bottom block, and the bottom block will be clear.\\

\noindent \textbf{Goal example}\\
b1 is on b2., b2 is on b3.\\

\noindent \textbf{Observation example}\\
b1 is on the table.  b2 is on the table.  B3 is on the table. Robot arm is empty. The b1 is clear. The b2 is clear. The b3 is clear. \\

\noindent \textbf{Action example}\\
pickup b2. 
    
\subsection{Gripper}
\textbf{Action List}

\noindent 1. move <room1> <room2>: This action allows the robot to move from one room to another.The action has a single precondition, which is that the robot is currently in a room. The effect of this action is to move the robot to another room and to remove the fact that it is in the original room.\\
2. pick <obj> <room> <gripper>: This action allows the robot to pick up an object using the gripper. The action has three preconditions: (1) the object is located in a room (2) the robot is currently in the same room and (3) the gripper is free (i.e., not holding any object). The effect of this action is to update the state of the world to show that the robot is carrying the object using the gripper, the object is no longer in the room, and the gripper is no longer free.\\
3. drop <obj> <room> <gripper>: This action allows the robot to drop an object that it is carrying. The action has two preconditions: (1) the robot is currently carrying the object using the gripper, and (2) the robot is currently in a room. The effect of this action is to update the state of the world to show that the robot is no longer carrying the object using the gripper, the object is now located in the room, and the gripper is now free.\\

\noindent \textbf{Goal example}\\
ball1 is at roomb. , ball2 is at roomb. , ball3 is at roomb. , ball4 is at room. \\

\noindent \textbf{Observation example}\\
Ball1 is a ball.  Ball1 is carrying right.  Ball2 is a ball.  Ball2 is at rooma.  Ball3 is a ball.  Ball3 is at rooma.  Ball4 is a ball.  Ball4 is at rooma.  Left is a gripper.  Left is free.  Right is a gripper.  Robby is at rooma.  Room rooma Room roomb.\\

\noindent \textbf{Action example}\\
Pick up ball1 at rooma with arm right.\\

\subsection{Tyreworld}

\noindent \textbf{Action List}\\
1. open <container>: The precondition for this action is that the container is unlocked and closed. The effect of this action is that the container is open and not closed. \\
2. close <container>: The precondition for this action is that the container is open. The effect of this action is that the container is closed and not open. \\
3. fetch <object> <container>: The precondition for this action is that the object is inside the container and the container is open. The effect of this action is that the object is held by the agent and not inside the container. \\
4. put-away <object> <container>: The precondition for this action is that the object is held by the agent and the container is open. The effect of this action is that the object is inside the container and not held by the agent. \\
5. loosen <nut> <hub>: The precondition for this action is that the agent has a wrench, the nut on hub is tight, and the hub is on the ground. The effect of this action is that the nut on hub is loose and not tight. \\
6. tighten <nut> <hub>: The precondition for this action is that the agent has a wrench, the nut on hub is loose, and the hub is on the ground. The effect of this action is that the nut on hub is tight and not loose. \\
7. jack-up <hub>: This action represents the process of lifting a hub off the ground using a jack. It requires the agent to have a jack and for the hub to be on the ground. After performing this action, the hub will no longer be on the ground and the agent will no longer have the jack. \\
8. jack-down <hub>: This action represents the process of lowering a hub back to the ground from an elevated position using a jack. It requires the agent to have the hub off the ground. After performing this action, the hub will be back on the ground and the agent will have the jack. \\
9. undo <nut> <hub>: This action undo the fastening of a nut on a hub. The preconditions are the hub is not on the ground (i.e., it has been jacked up), the hub is fastened, the agent has a wrench and the nut is loose. The effects are the agent has the nut, the hub is unfastened, the hub is no longer loose and the hub is not fastened anymore.\\ 
10. do-up <nut> <hub>: This action fasten a nut on a hub. The preconditions are the agent has a wrench, the hub is unfastened, the hub is not on the ground (i.e., it has been jacked up) and the agent has the nut to be fastened. The effects are the nut is now loose on the hub, the hub is fastened, the hub is no longer unfastened and the agent no longer has the nut. \\
11. remove-wheel <wheel> <hub>: This action removes a wheel from a hub. It can only be performed if the hub is not on the ground, the wheel is currently on the hub, and the hub is unfastened. After the action is performed, the agent will have the removed wheel and the hub will be free, meaning that the wheel is no longer on the hub. \\
12. put-on-wheel <wheel> <hub>: This action puts a wheel onto a hub. It can only be performed if the agent has the wheel, the hub is free, the hub is unfastened, and the hub is not on the ground. After the action is performed, the wheel will be on the hub, the hub will no longer be free, and the agent will no longer have the wheel. \\
13. inflate <wheel>: This action inflates a wheel using a pump. It can only be performed if the agent has a pump, the wheel is not inflated, and the wheel is intact. After the action is performed, the wheel will be inflated.\\

\noindent \textbf{Goal example}\\
w1 is in boot.\\

\noindent \textbf{Observation example}\\
Boot is closed. Boot is unlocked. Hub the-hub1 is fastened. Hub the-hub1 is on the ground. Jack is in boot. Pump is in boot. R1 is in boot. The nut nuts1 on the hub the-hub1 is tight. Wheel r1 is intact. Wheel r1 is not inflated. Wheel w1 is on hub the-hub1. Wrench is in boot.\\

\noindent \textbf{Action example}\\
Open boot.\\

\subsection{Barman}
\textbf{Action List}\\
1. <hand> grasp <container>: Grasp a container\\
2. <hand> leave <container>: Leave a container on the table\\
3. fill-shot <shot> <ingredient> <hand1> <hand2> <dispenser>: Fill a shot glass with an ingredient from dispenser\\
4. refill-shot <shot> <ingredient> <hand1> <hand2> <dispenser>: Refill a shot glass with an ingredient from dispenser\\
5. empty-shot <hand> <shot> <beverage>: Empty a shot glass
6. clean-shot <shot> <beverage> <hand1> <hand2>: Clean a shot glass\\
7. pour-shot-to-clean-shaker <shot> <ingredient> <shaker> <hand1> <level1> <level2>: Pour an ingredient from a shot glass to a clean shaker from level1 to level2\\
8. pour-shot-to-used-shaker <shot> <ingredient> <shaker> <hand1> <level1> <level2>: Pour an ingredient from a shot glass to a used shaker from level1 to level2\\
9. empty-shaker <hand> <shaker> <cocktail> <level1> <level2>: Empty a shaker containing cocktail from level1 to level2\\
10. clean-shaker <hand1> <hand2> <shaker>: Clean a shaker
11. shake <cocktail> <ingredient1> <ingredient2> <shaker> <hand1> <hand2>: Shake a cocktail in a shaker\\
12. pour-shaker-to-shot <beverage> <shot> <hand> <shaker> <level1> <level2>: Pour a beverage from a shaker to a shot glass from level1 to level2\\

\noindent \textbf{Goal example}\\
shot1 contains cocktail1.\\

\noindent \textbf{Observation example}\\
Cocktail1 part1 ingredient is ingredient1. Cocktail1 part2 ingredient is ingredient3. Cocktail2 part1 ingredient is ingredient2. Cocktail2 part2 ingredient is ingredient3. Cocktail3 part1 ingredient is ingredient1. Cocktail3 part2 ingredient is ingredient2. Dispenser1 dispenses ingredient1. Dispenser2 dispenses ingredient2. Dispenser3 dispenses ingredient3. Left hand is empty. Level l0 is next to level l1. Level l1 is next to level l2. Right hand is empty. Shaker1 is at empty level l0. Shaker1 is at level l0. Shaker1 is clean. Shaker1 is empty. Shaker1 is on the table. Shot1 is clean. Shot1 is empty. Shot1 is on the table. Shot2 is clean. Shot2 is empty. Shot2 is on the table. Shot3 is clean. Shot3 is empty. Shot3 is on the table. Shot4 is clean. Shot4 is empty. Shot4 is on the table.\\

\noindent \textbf{Action example}\\
right grasp shot1.\\

\subsection{Jericho}
\textbf{Action List} \\
1. Inventory: check things you are carrying\\
2. Look: check your surroundings\\
3. Examine <place/obj>: check the details of something\\
4. Take <obj>: pickup obj\\
5. Put down <obj>: leave a obj at your current place.\\
6. Drop <obj>\\
7. Check valid actions: Check actions you can use\\
8. South: go south\\
9. North: go north\\
10. East: go east\\
11. West: go west\\
12. Up: go up\\
13. Down: go down\\
14. Check valid actions (Other available actions)\\

\noindent \textbf{Goal example}\\
You are the warrior Link that needs to save the princess from the castle.\\

\noindent \textbf{Observation example}\\
You are at the path leading to the castle. The castle is to your north. There is a barrel in front of you.\\

\noindent \textbf{Action example}\\
Examine barrel\\

\section{Prompt Examples}
\label{appendix:prompt_examples}

\subsection{\std}

\begin{tcolorbox}[colback=gray!10!white,colframe=gray!50!white,title=Environment Implementation, sharp corners=south]
Your goal is to replace flat tyres with intact tyres on the hubs. Remember to open boot first to get tools you need. Intact tyres should be inflated. The nuts should be tight on the hubs. The flat tyres, wrench, jack, and pump should be in the boot. The boot should be closed.\\\\
    There are 13 actions defined in this domain:\\
    open <container>: The precondition for this action is that the container is unlocked and closed. The effect of this action is that the container is open and not closed.\\
    close <container>: The precondition for this action is that the container is open. The effect of this action is that the container is closed and not open.\\
    fetch <object> <container>: The precondition for this action is that the object is inside the container and the container is open. The effect of this action is that the object is held by the agent and not inside the container.\\
    put-away <object> <container>: The precondition for this action is that the object is held by the agent and the container is open. The effect of this action is that the object is inside the container and not held by the agent.\\
    loosen <nut> <hub>: The precondition for this action is that the agent has a wrench, the nut on hub is tight, and the hub is on the ground. The effect of this action is that the nut on hub is loose and not tight.\\
    tighten <nut> <hub>: The precondition for this action is that the agent has a wrench, the nut on hub is loose, and the hub is on the ground. The effect of this action is that the nut on hub is tight and not loose.\\
    jack-up <hub>: This action represents the process of lifting a hub off the ground using a jack. It requires the agent to have a jack and for the hub to be on the ground. After performing this action, the hub will no longer be on the ground and the agent will no longer have the jack.\\
    jack-down <hub>: This action represents the process of lowering a hub back to the ground from an elevated position using a jack. It requires the agent to have the hub off the ground. After performing this action, the hub will be back on the ground and the agent will have the jack.\\
    undo <nut> <hub>: This action undo the fastening of a nut on a hub. The preconditions are the hub is not on the ground (i.e., it has been jacked up), the hub is fastened, the agent has a wrench and the nut is loose. The effects are the agent has the nut, the hub is unfastened, the hub is no longer loose and the hub is not fastened anymore.\\
    do-up <nut> <hub>: This action fasten a nut on a hub. The preconditions are the agent has a wrench, the hub is unfastened, the hub is not on the ground (i.e., it has been jacked up) and the agent has the nut to be fastened. The effects are the nut is now loose on the hub, the hub is fastened, the hub is no longer unfastened and the agent no longer has the nut.\\
    remove-wheel <wheel> <hub>: This action removes a wheel from a hub. It can only be performed if the hub is not on the ground, the wheel is currently on the hub, and the hub is unfastened. After the action is performed, the agent will have the removed wheel and the hub will be free, meaning that the wheel is no longer on the hub.\\
    put-on-wheel <wheel> <hub>: This action puts a wheel onto a hub. It can only be performed if the agent has the wheel, the hub is free, the hub is unfastened, and the hub is not on the ground. After the action is performed, the wheel will be on the hub, the hub will no longer be free, and the agent will no longer have the wheel.\\
    inflate <wheel>: This action inflates a wheel using a pump. It can only be performed if the agent has a pump, the wheel is not inflated, and the wheel is intact. After the action is performed, the wheel will be inflated.\\\\
Here are examples:\\\\
The goal is to satisfy the following conditions: w1 is in boot. (Note you need to open boot first so that you can extract tools from it.)\\\\
Observation: Boot is closed. Boot is unlocked. Hub the-hub1 is fastened. Hub the-hub1 is on the ground. Jack is in boot. Pump is in boot. R1 is in boot. The nut nuts1 on the hub the-hub1 is tight. Wheel r1 is intact. Wheel r1 is not inflated. Wheel w1 is on hub the-hub1. Wrench is in boot.\\\\
Action: check valid actions.\\
Observation: Valid actions are: Open boot. Check valid actions. Look around.\\
Action: Open boot.\\
Observation: Boot is open. \\
Action: fetch wrench from boot\\
Observation: You have wrench.\\
Action: check valid actions\\
Observation: Valid actions are: Fetch pump from boot., Fetch r1 from boot., Put-away wrench in boot., Close boot., Fetch jack from boot., Loosen the nut nuts1 on the hub the-hub1., check valid actions\\
Action: Loosen the nut nuts1 on the hub the-hub1.\\
Observation: The nut nuts1 on the hub the-hub1 is loose. \\
Action: fetch jack from boot\\
Observation: You have jack. \\
Action: jack-up the hub the-hub1\\
Observation: Hub the-hub1 is not on the ground. \\
Action: Undo the fastening of the nut nuts1 on the hub the-hub1.\\
Observation: Hub the-hub1 is unfastened. \\
Action: Remove-wheel w1 from the hub the-hub1.\\
Observation: You have w1. \\
Action: Put-away w1 in boot\\
Observation: W1 is in boot. Goal is completed.\\
You should perform actions to accomplish the goal: The goal is to satisfy the following conditions: Wheel r1 is inflated., r1 is on the-hub1., w1 is in boot.\\
You should use the following commands for help when your action cannot be understood: check valid actions\\
Observation: Boot is closed. Boot is unlocked. Hub the-hub1 is fastened. Hub the-hub1 is on the ground. Jack is in boot. Pump is in boot. R1 is in boot. The nut nuts1 on the hub the-hub1 is tight. W1 is on the-hub1. Wheel r1 is intact. Wheel r1 is not inflated. Wrench is in boot.\\
Action: Open boot.\\
Observation: Boot is open.\\
Action: jack-up the-hub1\\
Observation: The action is not valid and therefore takes no effect. Please check valid actions.
Action: jack-up the hub the-hub1\\
Observation: The action is not valid and therefore takes no effect. Please check valid actions.\\
Action: Loosen the nut nuts1 on the hub the-hub1.\\
Observation: The action is not valid and therefore takes no effect. Please check valid actions.\\
Action: Loosen the nut nuts1 on the hub the-hub1.\\
Observation: The action is not valid and therefore takes no effect. Please check valid actions.\\
Action: Loosen the nut nuts1 on the hub the-hub1.\\
Observation: The action is not valid and therefore takes no effect. Please check valid actions.\\
Action: Fetch wrench from boot.\\
Observation: You have wrench.\\
\end{tcolorbox}

\subsection{\ours}

\begin{tcolorbox}[colback=gray!10!white,colframe=gray!50!white,title=Environment Implementation, sharp corners=south]
Your goal is to replace flat tyres with intact tyres on the hubs. Remember to open boot first to get tools you need. Intact tyres should be inflated. The nuts should be tight on the hubs. The flat tyres, wrench, jack, and pump should be in the boot. The boot should be closed.\\\\
    There are 13 actions defined in this domain:\\
    open <container>: The precondition for this action is that the container is unlocked and closed. The effect of this action is that the container is open and not closed.\\
    close <container>: The precondition for this action is that the container is open. The effect of this action is that the container is closed and not open.\\
    fetch <object> <container>: The precondition for this action is that the object is inside the container and the container is open. The effect of this action is that the object is held by the agent and not inside the container.\\
    put-away <object> <container>: The precondition for this action is that the object is held by the agent and the container is open. The effect of this action is that the object is inside the container and not held by the agent.\\
    loosen <nut> <hub>: The precondition for this action is that the agent has a wrench, the nut on hub is tight, and the hub is on the ground. The effect of this action is that the nut on hub is loose and not tight.\\
    tighten <nut> <hub>: The precondition for this action is that the agent has a wrench, the nut on hub is loose, and the hub is on the ground. The effect of this action is that the nut on hub is tight and not loose.\\
    jack-up <hub>: This action represents the process of lifting a hub off the ground using a jack. It requires the agent to have a jack and for the hub to be on the ground. After performing this action, the hub will no longer be on the ground and the agent will no longer have the jack.\\
    jack-down <hub>: This action represents the process of lowering a hub back to the ground from an elevated position using a jack. It requires the agent to have the hub off the ground. After performing this action, the hub will be back on the ground and the agent will have the jack.\\
    undo <nut> <hub>: This action undo the fastening of a nut on a hub. The preconditions are the hub is not on the ground (i.e., it has been jacked up), the hub is fastened, the agent has a wrench and the nut is loose. The effects are the agent has the nut, the hub is unfastened, the hub is no longer loose and the hub is not fastened anymore.\\
    do-up <nut> <hub>: This action fasten a nut on a hub. The preconditions are the agent has a wrench, the hub is unfastened, the hub is not on the ground (i.e., it has been jacked up) and the agent has the nut to be fastened. The effects are the nut is now loose on the hub, the hub is fastened, the hub is no longer unfastened and the agent no longer has the nut.\\
    remove-wheel <wheel> <hub>: This action removes a wheel from a hub. It can only be performed if the hub is not on the ground, the wheel is currently on the hub, and the hub is unfastened. After the action is performed, the agent will have the removed wheel and the hub will be free, meaning that the wheel is no longer on the hub.\\
    put-on-wheel <wheel> <hub>: This action puts a wheel onto a hub. It can only be performed if the agent has the wheel, the hub is free, the hub is unfastened, and the hub is not on the ground. After the action is performed, the wheel will be on the hub, the hub will no longer be free, and the agent will no longer have the wheel.\\
    inflate <wheel>: This action inflates a wheel using a pump. It can only be performed if the agent has a pump, the wheel is not inflated, and the wheel is intact. After the action is performed, the wheel will be inflated.\\\\
Note: A subgoal is a milestone goal that you need to complete in order to achieve the final goal. When there is an unfinished subgoal, you need to ground the given subgoal to corresponding executable actions for solving the given task in the following format: "Action: {action}". When there is no current subgoal or you believe the previous subgoal has been completed (based on past actions and observations), you need to output the next subgoal to be completed and its first action in the following format: "Subgoal: {subgoal}
Action: {action}". You cannot output two subgoals consecutively. Detailed trajectory information (action-observation pair) of previously satisfied subgoals will be hidden for context efficiency. If you believe that the detailed trajectory information of a particular subgoal is crucial for the current subgoal, you can use Action: "retrieve(subgoal\_id)" to obtain the detailed trajectory information.\\

Here are examples:\\\\
The goal is to satisfy the following conditions: w1 is in boot. (Note you need to open boot first so that you can extract tools from it.)\\\\
Observation: Boot is closed. Boot is unlocked. Hub the-hub1 is fastened. Hub the-hub1 is on the ground. Jack is in boot. Pump is in boot. R1 is in boot. The nut nuts1 on the hub the-hub1 is tight. Wheel r1 is intact. Wheel r1 is not inflated. Wheel w1 is on hub the-hub1. Wrench is in boot.\\\\
1 Subgoal: Open the boot to access the tools.\\
Action: check valid actions.\\
Observation: Valid actions are: Open boot. Check valid actions. Look around.\\
Action: Open boot.\\
Observation: Boot is open. Subgoal is met.\\\\
2 Subgoal: Fetch the wrench from the boot.\\
Action: fetch wrench from boot\\
Observation: You have wrench. Subgoal is met. \\\\
3 Subgoal: Loosen the nut on the hub.\\
Action: Loosen the nut nuts2 on the hub the-hub1.\\
Observation: The action is not valid. Subgoal is not met.\\\\
4 Subgoal: Examine errors and try to loosen the nut on the hub again.\\
Action: retrieve(3)\\
Observation: trajectory of Subgoal 3 is retrieved.\\
Action: check valid actions\\
Observation: Valid actions are: Fetch pump from boot., Fetch r1 from boot., Put-away wrench in boot., Close boot., Fetch jack from boot., Loosen the nut nuts1 on the hub the-hub1., check valid actions\\
Action: Loosen the nut nuts on the hub the-hub1.\\
Observation: The nut nuts1 on the hub the-hub1 is loose. Subgoal is met.\\\\
5 Subgoal: Fetch the jack from the boot and jack up the hub.
Action: fetch jack from boot\\
Observation: You have jack.\\
Action: jack-up the hub the-hub1\\
Observation: Hub the-hub1 is not on the ground. Subgoal is met.\\\\
6 Subgoal: Unfasten the nut nuts1 and remove wheel w1.\\
Action: Undo the fastening of the nut nuts1 on the hub the-hub1.\\
Observation: Hub the-hub1 is unfastened.\\
Action: Remove-wheel w1 from the hub the-hub1.\\
Observation: You have w1. Subgoal is met.\\\\
7 Subgoal: Put away the wheel w1 in the boot.\\
Action: Put-away w1 in boot\\
Observation: W1 is in boot. Goal is completed.
You should perform actions to accomplish the goal: The goal is to satisfy the following conditions: Wheel r1 is inflated., r1 is on the-hub1., w1 is in boot.\\\\
You should use the following commands for help when your action cannot be understood: check valid actions\\\\
Observation: Boot is closed. Boot is unlocked. Hub the-hub1 is fastened. Hub the-hub1 is on the ground. Jack is in boot. Pump is in boot. R1 is in boot. The nut nuts1 on the hub the-hub1 is tight. W1 is on the-hub1. Wheel r1 is intact. Wheel r1 is not inflated. Wrench is in boot.\\\\
1 Subgoal:  Open the boot to access the tools.\\
Observation: Boot is open.\\\\
2 Subgoal:  Fetch the wrench from the boot.\\
Observation: You have wrench.\\\\
3 Subgoal:  Loosen the nut on the hub the-hub1.\\
Observation: The nut nuts1 on the hub the-hub1 is loose.\\\\
4 Subgoal:  Fetch the jack from the boot and jack up the hub the-hub1.\\
Observation: You have jack and hub the-hub1 is elevated. Subgoal is met.\\\\
5 Subgoal:  Unfasten the nut nuts1 and remove wheel w1 from the hub the-hub1.\\
Action: undo nuts1 on the-hub1\\
Observation: Hub the-hub1 is unfastened. You have nuts1.\\
Action: Remove-wheel w1 from the-hub1\\
Observation: The-hub1 is free.  You have w1.

\end{tcolorbox}

\end{document}